\documentclass[10pt, conference, compsocconf]{IEEEtran}
\usepackage[bookmarks=false]{hyperref}

\usepackage{graphicx}

\ifCLASSINFOpdf
\else
\fi
\usepackage{pdfcomment}

\begin{document}
%
\title{Automatic Photo Orientation Detection with Convolutional Neural Networks}


\author{\IEEEauthorblockN{Ujash Joshi and Michael Guerzhoy}
\IEEEauthorblockA{Dept. of Computer Science\\
University of Toronto\\
Toronto, Ontario, Canada\\
ujash.joshi@utoronto.ca, guerzhoy@cs.toronto.edu}}


%


\maketitle
\begin{abstract}
We apply convolutional neural networks (CNN) to the problem of image orientation detection in the context of determining the correct orientation (from 0, 90, 180, and 270 degrees) of a consumer photo. The problem is especially important for digitazing analog photographs. We substantially improve on the published state of the art in terms of the performance on one of the standard datasets, and test our system on a more difficult large dataset of consumer photos. We use Guided Backpropagation to obtain insights into how our CNN detects photo orientation, and to explain its mistakes.

\end{abstract}


%
\IEEEpeerreviewmaketitle

\section{Introduction}

\begin{figure}[b]
\centering
\begin{tabular}{llll}
\includegraphics[height=40pt]{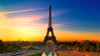} & \includegraphics[height=40pt]{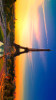} & \includegraphics[height=40pt]{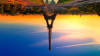} & \includegraphics[height=40pt]{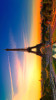} \\
$0^{\circ}$ & $90^{\circ}$ & $180^{\circ}$ & $270^{\circ}$
\end{tabular}
\caption{Correct outputs for different inputs. The possible outputs are $0^{\circ}$, $90^{\circ}$, $180^{\circ}$, and $270^{\circ}$.\label{fig:ori}}
\end{figure}

In this paper, we address the problem of detecting the correct orientation of a consumer photograph (i.e., $0^{\circ}$, $90^{\circ}$, $180^{\circ}$, or $270^{\circ}$; see Figure~\ref{fig:ori}) by learning a deep convolutional neural network (CNN). We experiment with standard datasets, on one of which our system performs substantially better than the published state of the art, and we experiment on a large dataset of consumer photos that we collected. We apply Guided Backpropagation~\cite{riedmiller2015striving}~\cite{zeiler2014visualizing} in order to visualize what our classifier is doing and to explain  the mistakes it makes. 

We detect the orientation of a photo by learning a classifier that classifies input images into four classes: $0^{\circ}$, $90^{\circ}$, $180^{\circ}$, or $270^{\circ}$. Our classifier is a deep convolutional neural network whose architecture is a modification of VGG-16~\cite{simonyan2014very}, a commonplace architecture used for image classification. We train our classifier on large datasets of photos.

Automatic photo orientation detection can help with speeding up the digitization of analog photos. It is a well-studied problem~\cite{vailaya2002automatic}. To date, learning-based approaches to the problem~\cite{vailaya2002automatic}~\cite{ciocca2015image}~\cite{wang2003image} consisted of exctracting low-level features used in image classification and retrieval such as Histograms of Gradients (HoG)~\cite{dalal2005histograms} and Colour Moments~\cite{vailaya1999content}, and sometimes high-level features such as face and object detector outputs~\cite{luo2005automatic}, and then feeding them into a learned classifier. Such classifiers perform very well on some standard datasets of photos. Examples of such datasets include the Corel stock photo dataset~\cite{duygulu2002object}, which consists of professional photos, and the SUN-497 database~\cite{xiao2010sun} where each photo is labelled as containing a particular scene. In recent years, convolutional neural networks have been used instead of classifying hand-engineered features in object recognition~\cite{krizhevsky2012imagenet},  image retrieval~\cite{babenko2014neural} and the estimation of image skew~\cite{fischer2015image} (note that this is a distinct problem from the one addressed in this paper: we are interested in accurately classifying a photo into four possible orientation bins, while Fischer et al. attempt to estimate the skew angle, which could be any real number.) In this work, we do the same for the related problem of photo orientation detection. Cao et al.~\cite{caofast2016} describe another biologically-inspired approach to the estimation of image skew, using a shallow architecture.

Recent visualization techniques for CNNs~\cite{riedmiller2015striving}~\cite{zeiler2014visualizing}~\cite{yosinski2015understanding} have mostly been used for visualizing the function of particular neurons in a deep neural network, but they also allow for exploring how and why CNNs classify images the way they do~\cite{selvaraju2016grad}. We use Guided Backpropagation in order to visualize how our network classifies and misclassifies the orientation of photos, and obtain insight into how it works. 

The rest of the paper is organized as follows. We outline our modifications to the VGG-16 architecture to obtain a photo orientation classifier, and detail our training procedure. We then present our experimental results on the standard datasets for the task of photo orientation detection, and compare them to prior work, demonstrating that CNNs are able to detect the orientation more accurately than prior work. We describe our own dataset of consumer photos, and analyze our experimental results on that dataset. Finally, we visualize what our CNN is doing in order to obtain insights into how CNNs detect photo orientation. Our contribution consists of obtaining better than published-state-of-the-art results on the task of image orientation detection, and a demonstration of the use of Guided Backpropagation for analyzing the outputs of a deep neural network.

\section{Modifying the VGG-16 architecture to build a photo orientation classifier}

A common technique for building a CNN classifier for a new domain is to adopt an architecture orginally designed for the ImageNet dataset~\cite{deng2009imagenet}, modify it, and apply it to the new domain. See e.g.~\cite{rothe2015dex} and~\cite{gomez2016animal}. We found that an architecture that is identical to VGG-16, except with $4$ outputs corresponding to $0^{\circ}$, $90^{\circ}$, $180^{\circ}$, or $270^{\circ}$ instead of $1,000$ outputs corresponding to the $1,000$ object classes in ImageNet, performed the best on our datasets.

\subsection{Training the CNN}

We found that initializing the weights of our network to the weights of VGG-16 trained on ImageNet, and then training the network end-to-end resulted in the best validation performance. This indicates that we are doing some transfer learning: VGG-16 detects 1000 classes of objects, which would be useful for detecting orientation. Likely initializing our weights to those of VGG-16 makes our network converge to nearby values of the weights.

The set of photos is transformed by rotating all the photos in the original training set by $0^{\circ}$, $90^{\circ}$, $180^{\circ}$, or $270^{\circ}$. The VGG architecture requires that the input be of size $224\times 224\times 3$. We resize the input image to fit inside a $224\times 224$ square, and pad it as necessary with black pixels in order for the input to be $224\times 224\time 3$.

The network is trained using Dropout with $p = .7.$

\section{Experimental results}
\subsection{Prior work}
Ciocca et al.~\cite{ciocca2015image} summarize the current state of the art in photo orientation detection on two standard datasets: the Corel stock photo dataset~\cite{duygulu2002object} and the SUN-397 database~\cite{xiao2010sun}. On the SUN-397 database, the best results were obtained by Ciocca et al.~\cite{ciocca2015image}, with 92.4\% accuracy. On the Corel dataset, the best results were obtained by Vailaya et al.~\cite{vailaya2002automatic}, with 97.4\% accuracy.

\subsection{Dataset descriptions \label{sec:datadesc}}

The Corel dataset consists fo approx. 10,000 images, separated into 80 concept groups such as autumn, aviation, bonsai, castle, and waterfall. The SUN database consists consists of about 108,000 images, separated into 397 categories. Our own dataset, collected from Flickr by downloading images corresponding to 26 tags, consists of about 250,000 images.

\begin{figure}
\centering
\begin{tabular}{lll}
\includegraphics[height=60pt]{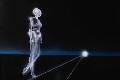} & \includegraphics[height=60pt]{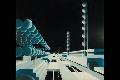} & \includegraphics[height=60pt]{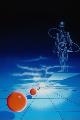}
\end{tabular}
\caption{Some images from the Corel dataset are not representative of consumer photos.\label{fig:corel_weird}}
\end{figure}

Some of the images in the Corel dataset have very low-resolution. They have been resized to be larger but to still fit into a $224\times 224$ square. Some images in the Corel dataset are atypical of consumer photos. Sample images from the \texttt{art\_cybr} category of the Corel dataset are shown in Fig.~\ref{fig:corel_weird}.

We split all datasets into training (64\%), test (20\%), and validation (16\%) sets, and then transform each of the sets by adding in all the possible rotations of each photo.

\subsection{Experimental results}

The accuracy of our classifiers on the test sets of the datasets under consideration are summarized in Table~\ref{tab:results}. The results for the Corel dataset should be interpreted with caution because of the issues described in Section~\ref{sec:datadesc}. We have matched or exceeded the published state of the art on both standard datasets for the task.

\begin{table}[]
\centering
\begin{tabular}{|l|l|l|}
\hline
Dataset 	& Accuracy (ours) & {Accuracy (SOTA)} \\ \hline
Flickr (ours)  		&  92.5\%                              &    					\\ \hline
SUN 397  	&  98.5\%                             &   		92.4\% (Ciocca et al., 2015\cite{ciocca2015image})			\\ \hline
Corel   		&   97.5\%                                       &             97.4\% (Vailaya et al., 2002\cite{vailaya2002automatic})                  \\ \hline
\end{tabular}
\caption{Experimental results for classifying the orientation of photos \label{tab:results}}

\end{table}

\subsection{Discussion}

Our results show that convolutional neural networks match or outperform the published state of the art in image orientation detection on both standard datasets. The Corel dataset appears to not be diverse enough: we suspect that we are overfitting on some of the categories -- we are including photos from all categories in both the training and the test set. Our results on our own Flickr dataset indicate that the SUN dataset may not be fully representative of consumer photos. This would not be an issue on our Flickr dataset, since all of our categories are ubiquitous in consumer photos and there is a large degree of intra-vategory diversity.

\section{Understanding the CNN photo orientation detector using visualization}

In this work, we have shown that a deep architecture is able to detect photo orientation better than any of the published results employing shallow architectures that use combinations of low and high level features. It is of interest to see how the deep architecture is able to classify the photos, both in order to understand how the deep archticture classifies the photos, and in order to explain its mistakes. We show how to use  Guided Backpropagation~\cite{riedmiller2015striving} to better understand what our CNN is doing. 

Visualizing CNNs involves visualizing the roles of individual neurons. To visualize the roles of an individual neuron, researchers found patches of real images that activate that neuron the most~\cite{zeiler2014visualizing}, used methods similar to gradient ascent in order to synthesize images that activate that neuron the most~\cite{yosinski2015understanding}, or visualized the change in images that would increase the activity of the neuron the most~\cite{riedmiller2015striving}~\cite{zeiler2014visualizing}. These approaches can also be used in combination with each other. Recent work~\cite{selvaraju2016grad} employed Guided Backpropagation in the context of object recognition.

We are interested, for every image in the test set, in explaining why our CNN obtained the answer that it did. That means that, when the input is a specific image of interest, we want to visualize the output neuron of our CNN whose activity is the largest of all four output neurons.

\subsection{Guided Backpropagation \label{sec:guidedbp}}
We use a variant of Guided Backpropagation to explain the activity of our output neurons. Guided Backpropagation computes a modified version of the gradient of a particular neuron with respect to the input. We display that modified gradient as a saliency map. We are interested in an explanation for the network's output. For that reason, if, for a specific image $x$, the network's maximal output is the $m$-th unit $p_m$, we produce a saliency map that is computed similarly to  $\partial p_m/\partial x$, but is clearer than the gradient.

If the absolute value of the gradient $|\frac{\partial neuron}{\partial x_i}|$ is large, that means that increasing (or decreasing) $x_i$ would influence the neuron. However, there can be a number of mechanisms for that to happen: one possibility is that the pixel $x_i$ currently activates a feature that, when activated, increases the activity of a higher level feature, which in turn activates an even higher-level feature, which in turn activates the neuron of interest. Another possibility is that the pixel $x_i$ activates a feature that in turn turns off a higher-level feature, which in turn activates an even higher-level feature, which in turn activates the neuron of interest. We do not want to visualize $x_i$ as influecing the output neuron in case $|\frac{\partial neuron}{\partial x_i}|$ is large for the second reason. That is because if $x_i$'s changing depresses some feature more, causing the final output to be higher, $x_i$ provides evidence for the \textit{absence} of some feature in the image. Since numerous features are absent but only a few are present, it makes less sense to take into account evidence for the absence of features when visualizing the saliency map that indicates which pixels influence the output. Empirically, $\frac{\partial neuron}{\partial x}$ is very noisy~\cite{riedmiller2015striving}.

\begin{figure}[h]
\begin{center}
\includegraphics[width=180pt]{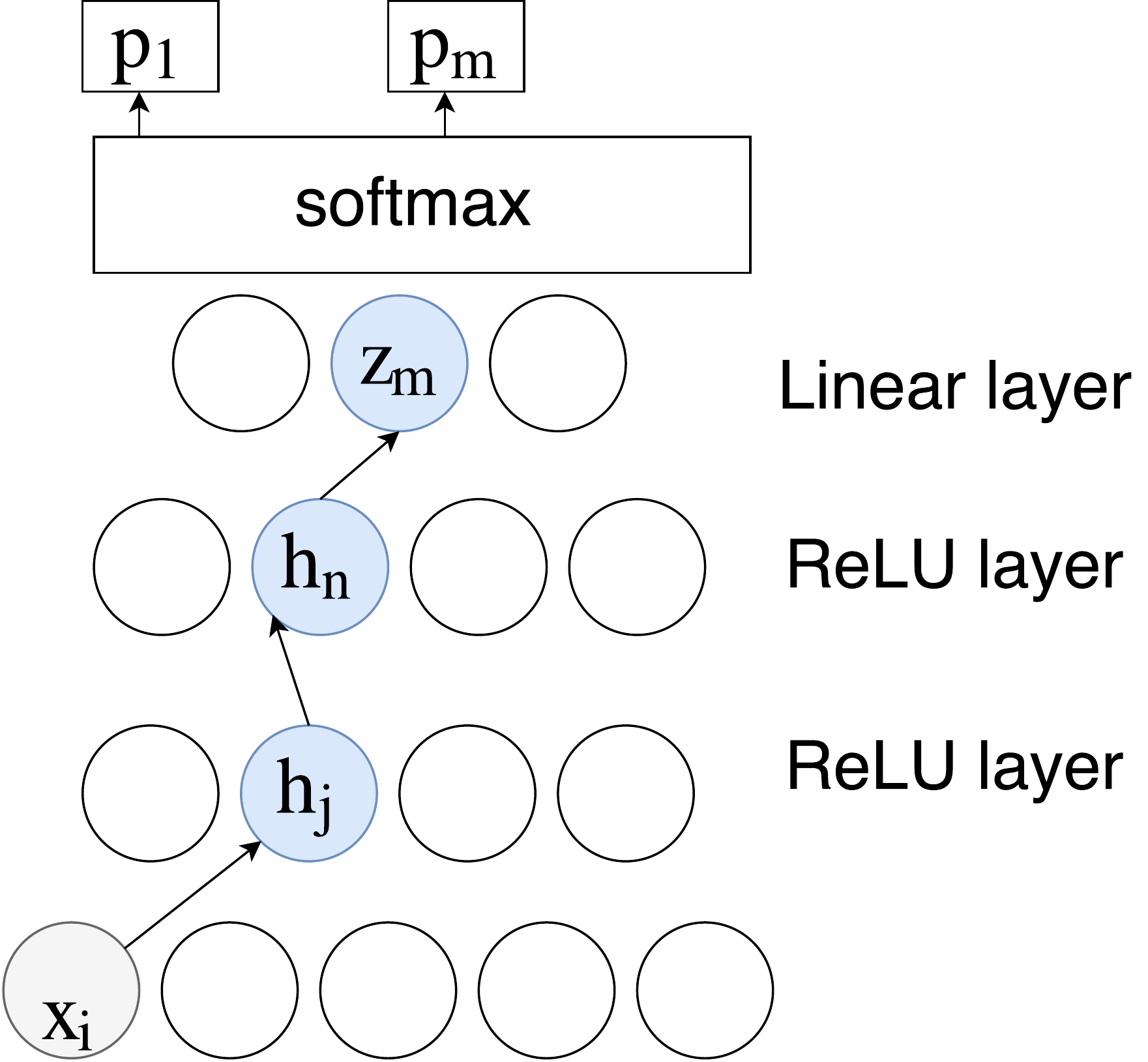}
\end{center}
\caption{A path in a network from $x_i$ to $p_m$ where all the units along the path are activated and the weights connecting them are all positive. $x_i$ would be visualized on the saliency map when using Guided Backpropagation. \label{fig:evidence_chain}}
\end{figure}

Guided backpropagation is a way of visualizing which pixels provide evidence for the \textit{presence} of features in the input image that influence the output neuron. The pixels that are visualized never \textit{depress} features causing the neuron of interest to activate. Instead, they only activate features throughout the layers of the network. This leads to much clearer visualizations. For the network in Fig.~\ref{fig:evidence_chain}, the pixel $x_i$ will be prominent in the saliency map that corresponds to the output $z_m$ only if there is a path between $x_i$ and $z_m$ such that all the hidden ReLU units along that path are activated and all the partial derivatives along that path (i.e., $\partial h_n/\partial h_j$, and $\partial h_j/\partial x_i$) are positive.

(Note that in a network that only uses ReLU activation functions, we can speak of features that correspond to ReLU units being turned ``activated" and ``depressed," referring to neurons' outputs' being positive or zero respectively. With activation functions that can take positive or negative values, this would not be possible.) 

\begin{figure*}[h]
\centering
\begin{tabular}{ll}
\includegraphics[width=180pt]{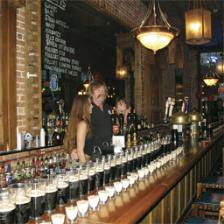} & \includegraphics[width=180pt]{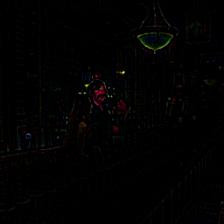}
\end{tabular}
\caption{A correctly-oriented photo. The Guided Backpropagation visualization indicates that the outline of the light fixture was a cue for correctly orienting the image.\label{fig:gg_light_correct}}
\end{figure*}

\begin{figure*}[h]
\centering
\begin{tabular}{ll}
\includegraphics[width=180pt]{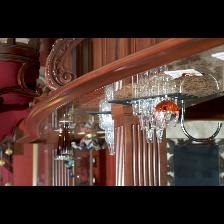} & \includegraphics[width=180pt]{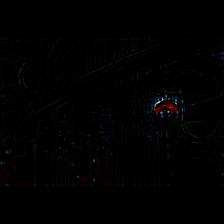}
\end{tabular}
\caption{The classifer output the photo is upright, but it should be rotated by  $180^{\circ}$. The Guided Backpropagation visualization indicates that the outline of the wine glass was useful for orienting the photo, suggesting that the wineglass was mistaken for a light fixture. \label{fig:gg_light_incorrect}}
\end{figure*}

\begin{figure*}[h]
\centering
\begin{tabular}{ll}
\includegraphics[width=180pt]{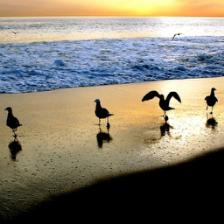} & \includegraphics[width=180pt]{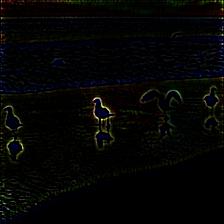}
\end{tabular}
\caption{A correctly-oriented photo. The Guided Backpropagation visualization indicates that the shapes of the birds were a cue for correctly orienting the image. \label{fig:gg_birds}}
\end{figure*}

\begin{figure*}[h]
\centering
\begin{tabular}{ll}
\includegraphics[width=180pt]{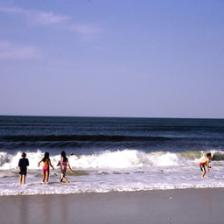} & \includegraphics[width=180pt]{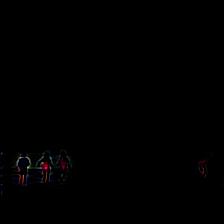}
\end{tabular}
\caption{A correctly-oriented photo. The Guided Backpropagation visualization indicates that people were a cue for correctly orienting the image. \label{fig:gg_birds}}
\end{figure*}

The saliency map that visualizes what a neuron of interest $p_m$ is doing for a specific image is computed using Guided Backpropgation as follows. Partial derivatives are computed as if a Backpropagation pass is being computed, except that negative partial derivatives are set to 0 before proceeding to the layer below each time.  The result is a ``modified version" of $\partial p_m/\partial x$. The modified $\partial p_m/\partial x$ is high for $x_i$'s that, if they are increased, increase the activations of already-active hidden neurons that correspond to features detected in the image that contribute to $p_m$'s being high.

The result is a saliency map where pixels that provide positive evidence for features that contribute to the output $z_m$'s being high are displayed. 

Most of the pixels on the computed saliency map are generally black. There are two reasons for this. First, for most $x_i$,  $\partial p_m/\partial x_i$ is very close to 0, since most pixels do not activate higher-level features. Second, since in the salienct map produced using Guided Backpropagation only pixels that provide positive evidence for $p_m$'s being high all they way up the network are displayed, there are many more 0-valued pixels in the saliency map than in $\partial p_m/\partial x$.

\subsection{Explaining correct predictions by the CNN using Guided Backpropagation}

In this section, we provide several examples of explanations of how the CNN detected the correct orientation of photos that were generated using Guided Backpropagation. The explanations are generated by computing the Guided Backpropagation saliency map using the algorithm described in Section~\ref{sec:guidedbp} with respect to the output $p_i$, where $i$ is the correct orientation, and $p_i$ was the largest output. The interpretations of the visualizations are necessarily speculative, but the visualizations are suggestive.

Light fixtures are usually reliable cues for orienting indoor photos. In Figure~\ref{fig:gg_light_correct}, we display an example of a correctly oriented indoor photo, together with a Guided Backpropagation visualization. Interestingly, it is the shape of the light fixture that seems to be the cue. Items that look like light fixture seem to sometimes mislead the classifier. For example, in Figure~\ref{fig:gg_light_incorrect}, it appears that a wine glass was ``mistaken" by the classifier for a light fixture.

Objects commonly found in scenes can be useful for orienting a photo. For example, in Figure~\ref{fig:gg_birds}, it appears that the shapes of the birds were useful in correctly orienting the photos.

\subsection{Explaining mistakes by the CNN using Guided Backpropagation}
In this section, we provide several examples of explanations of how the CNN detected the \textit{incorrect} orientation of a photo that were generated using Guided Backpropagation. One example (Figure~\ref{fig:gg_light_incorrect}) was already shown. It appears that the CNN detects numerous objects and uses object detections as cues for orientation detection. In the example in Figure~\ref{fig:gg_light_incorrect}, the CNN seems to have incorrectly identified a wineglass as a light fixture.

\begin{figure*}
\centering
\begin{tabular}{ll}
\includegraphics[width=180pt]{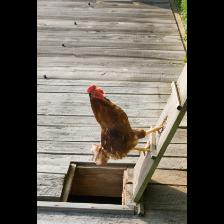} & \includegraphics[width=180pt]{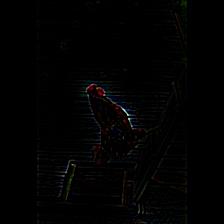}
\end{tabular}
\caption{An incorrectly-oriented photo. The classifier output suggested the photo is upright, but it should be rotated by  $90^{\circ}$. The Guided Backpropagation visualization indicates that the chicken was misdetected. \label{fig:gg_chicken}}
\end{figure*}

In Fig~\ref{fig:gg_chicken}, another interesting mistake is made. It appears that the rooster is used as a cue, but the image is nevertheless oriented incorrectly by the classifer. From the visualization, it appears plausible that the network would ``think" that the bird it detected is oriented upright in the incorrectly-rotated image.

\section{Conclusions and Future Work}
In this paper, we demonstrated that deep convolutional neural networks (CNN) outperform shallow architectures for the task of image orientation detection. We used Guided Backpropagation in order to explain both the correct and incorrect outputs of our classifier. We have shown that the CNN uses object detections in order to perform image orientation detection. Further evidence of this is that initializing the weights of our CNN to be the same as those in the VGG-16 network trained on ImageNet, suggesting that transfer learning is useful for image orientation detection (since it is likely that the we converge on weights that are close to the weights of VGG-16 for the lower layers if we initialize our weights to be those of VGG-16).

We plan to systematically study the outputs of our Guided Backpropagation visualizations in order to obtain quantitative insights about the behaviour of the CNN.




%
%
%

\bibliographystyle{IEEEtran}
\bibliography{IEEEabrv,imageOri}

\end{document}